\pdfoutput=1

\documentclass[11pt]{article}

\usepackage[preprint]{acl}

\usepackage{times}
\usepackage{latexsym}
\usepackage{amsmath}
\usepackage{amssymb}
\usepackage{txfonts}
\usepackage{graphicx}
\usepackage{booktabs}
\usepackage{subcaption}
\usepackage{enumitem}
\usepackage{makecell}

\usepackage[T1]{fontenc}

\usepackage[utf8]{inputenc}

\usepackage{microtype}

\usepackage{inconsolata}

\usepackage{graphicx}

\newcommand{\condprob}[2]{\mathsf{P}( #1 \mid #2 )}

\title{The Inverse Scaling Effect of Pre-Trained Language Model Surprisal \\ Is Not Due to Data Leakage}

\author{Byung-Doh Oh \\
  Center for Data Science \\
  New York University \\
  \texttt{oh.b@nyu.edu} \\\And
  Hongao Zhu \\
  School of Foreign Languages \\
  Shanghai Jiao Tong University \\
  \texttt{cisco\_sfl\_sjtu@sjtu.edu.cn} \\\And
  William Schuler \\
  Department of Linguistics \\
  The Ohio State University \\
  \texttt{schuler.77@osu.edu}}

\begin{document}
\maketitle
\begin{abstract}
In psycholinguistic modeling, surprisal from larger pre-trained language models has been shown to be a poorer predictor of naturalistic human reading times.
However, it has been speculated that this may be due to data leakage that caused language models to see the text stimuli during training.
This paper presents two studies to address this concern at scale.
The first study reveals relatively little leakage of five naturalistic reading time corpora in two pre-training datasets in terms of length and frequency of token $n$-gram overlap.
The second study replicates the negative relationship between language model size and the fit of surprisal to reading times using models trained on `leakage-free' data that overlaps only minimally with the reading time corpora.
Taken together, this suggests that previous results using language models trained on these corpora are not driven by the effects of data leakage.
\end{abstract}

\section{Introduction}
Language models (LMs) based on neural networks, which are trained to predict upcoming words, have been shown to flexibly capture many linguistic regularities from raw text \citep{linzenbaroni21, mahowaldetal24}.
This has sparked research at the intersection between language modeling and psycholinguistics that relates LM probabilities to human behavior.
One line of such research focuses on evaluating LM surprisal \citep[negative log probabilities;][]{shannon48} against measures of processing difficulty such as word-by-word reading times, under an `expectation-based' theoretical link that posits predictability as a key determinant of processing difficulty \citep{hale01, levy08}.

However, the source stimuli of the reading time datasets \citep[e.g.][]{futrelletal21, lukechristianson18} used in such studies are often naturalistic text that are available online (e.g.~news articles), which raises the concern that those texts may occur in the LMs' pre-training corpora.
If the degree of such data leakage is severe, the LMs may assign artificially lower surprisal to the text in reading time datasets as a result of having `memorized' it during training.
As a consequence, this could bring into question the validity of previous results as well as the general practice of using pre-trained LMs in psycholinguistic modeling.
For example, it has been speculated that the negative relationship between the size of an LM and the fit of its surprisal to human reading times observed on English data \citep[e.g.][]{ohschuler23tacl, shainetal24} may be due to such leakage \citep{wilcoxetal23qp}.

This work presents two studies to address this concern at scale.
The first study assesses the leakage of five naturalistic reading time corpora in two pre-training datasets that were each used to train Pythia and GPT-2 LMs \citep{bidermanetal23, radfordetal19} by identifying the longest overlapping token sequence and its frequency.
The second study then uses the same methodology to curate training data that overlaps minimally with the reading time corpora, and trains LMs on it to examine whether the negative relationship between model size and the fit of LM surprisal is observed with `leakage-free' training data.
Additionally, data leakage is artificially introduced through fine-tuning to study how LM surprisal's fit to reading times would change in the face of severe leakage.

The results indicate that commonly used reading time corpora suffer little from data leakage, with most passages sharing only relatively short overlaps among the billions of tokens in the two pre-training corpora.
Moreover, LMs trained on leakage-free data robustly replicate the negative relationship between model size and surprisal's fit to reading times, further indicating that this phenomenon is not simply due to leakage.
However, results also show that actual severe leakage is likely to result in an overestimation of this negative relationship, which still warrants caution against the leakage of reading time corpora.
Taken together, these results suggest that previous findings based on LMs trained on these corpora are not due to the effects of data leakage.

\section{Study 1: Overlap Between Reading Time and Pre-Training Corpora} \label{sec:exp1}
The first study assesses the leakage of naturalistic reading time corpora in LM pre-training datasets.
To this end, Compacted Directed Acyclic Word Graphs \citep[CDAWGs;][]{crochmoreverin97, inenagaetal05} were built on two pre-training corpora, which allows the reading time corpora to be queried efficiently to identify the longest overlapping token sequence and its frequency.

\subsection{Methods} \label{sec:exp1_methods}
\paragraph{Pre-Training Corpora.} The two English pre-training corpora that were analyzed in this study are the subset of the Pile \citep{gaoetal20} that was used to train the Pythia LMs \citep{bidermanetal23}, and the OpenWebText Corpus \citep{gokaslancohen19}, which is an open-source replication of the GPT-2 LMs' \citep{radfordetal19} training data.
The training data of the Pythia LMs is provided as pre-tokenized examples of length 2,049, which are sequences sampled from a concatenated version of the Pile.
A total of 143,000 batches that each contain 1,024 training examples were used to train the Pythia LMs, which amounts to a total of $\sim$300B tokens.
The OpenWebText Corpus consists of 8,013,769 documents, which is equivalent to $\sim$8.7B tokens when tokenized with Pythia LM's subword tokenizer.
\paragraph{Reading Time Corpora.} The five English reading time corpora that served as queries are:
\begin{enumerate}[leftmargin=*, itemsep=0pt]
    \item Dundee \citep{kennedyetal03}: 67 newspaper editorials from \textit{The Independent}.
    \item Brown \citep{smithlevy13}: 13 passages from the Brown Corpus \citep{kucerafrancis67}.
    \item GECO \citep{copetal17}: 13 chapters from the novel \textit{The Mysterious Affair at Styles} \citep{christie20}.
    \item Provo \citep{lukechristianson18}: 55 passages of news articles, science magazine excerpts, and fictional work.
    \item Natural Stories \citep{futrelletal21}: 10 passages of narrative and expository text.
\end{enumerate}

With the exception of the Dundee Corpus, most of the source text in these corpora are available online, which makes them susceptible to leakage in pre-training corpora.
Additionally, while Natural Stories has been manually edited to include challenging syntactic constructions, there is still likely to be substantial overlap if the pre-training corpora contain the original source text.

\paragraph{CDAWG Construction and Querying.} A CDAWG is a finite-state machine that is specialized for indexing sequences, which allows the length of the longest occurring suffix of a query to be returned efficiently.
We use \citeauthor{merrilletal24}'s \citeyearpar{merrilletal24} implementation\footnote{\url{https://github.com/viking-sudo-rm/rusty-dawg}} to build CDAWGs on the two pre-training corpora after normalizing their line breaks and whitespaces and tokenizing them with Pythia LM's subword tokenizer.
Training examples from the Pile were additionally split at \texttt{<|endoftext|>} tokens in order to treat text from different documents as separate sequences.

Subsequently, each passage of the five reading time corpora was tokenized using the same tokenizer and queried against the two CDAWGs.
The length of the longest occurring suffix was then calculated at every token position to retrieve the globally longest overlapping token sequence between each passage and the pre-training corpora.\footnote{Borrowing the example of \citet{merrilletal24}, querying \texttt{l l o y d} against the reference \texttt{h e l l o w o r l d} returns the lengths <1, 2, 3, 0, 1> at each token position, which allows the longest overlapping sequence \texttt{l l o} to be identified at token position 3.}
The frequency of this sequence in the pre-training corpora was also retrieved to further gauge the severity of this overlap.
Additionally, the joint probability of this sequence was calculated using a 5-gram LM with backoff using the KenLM toolkit \citep{heafieldetal13} with parameters estimated on the Gigaword 4 corpus \citep{parkeretal09}.
This allows us to identify sequences that are likely to appear in a corpus of a given size at roughly chance level.\footnote{The probability of the sequence appearing at least once was estimated as $1-(1-p)^n$, where $p$ is the probability of the sequence and $n$ is the number of whitespace words in each corpus. The $p$ that sets the probability of this event to some threshold can then be calculated for each pre-training corpus.}

\begin{figure*}[ht!]
    \centering
    \begin{subfigure}[t]{\textwidth}
        \includegraphics[width=\textwidth]{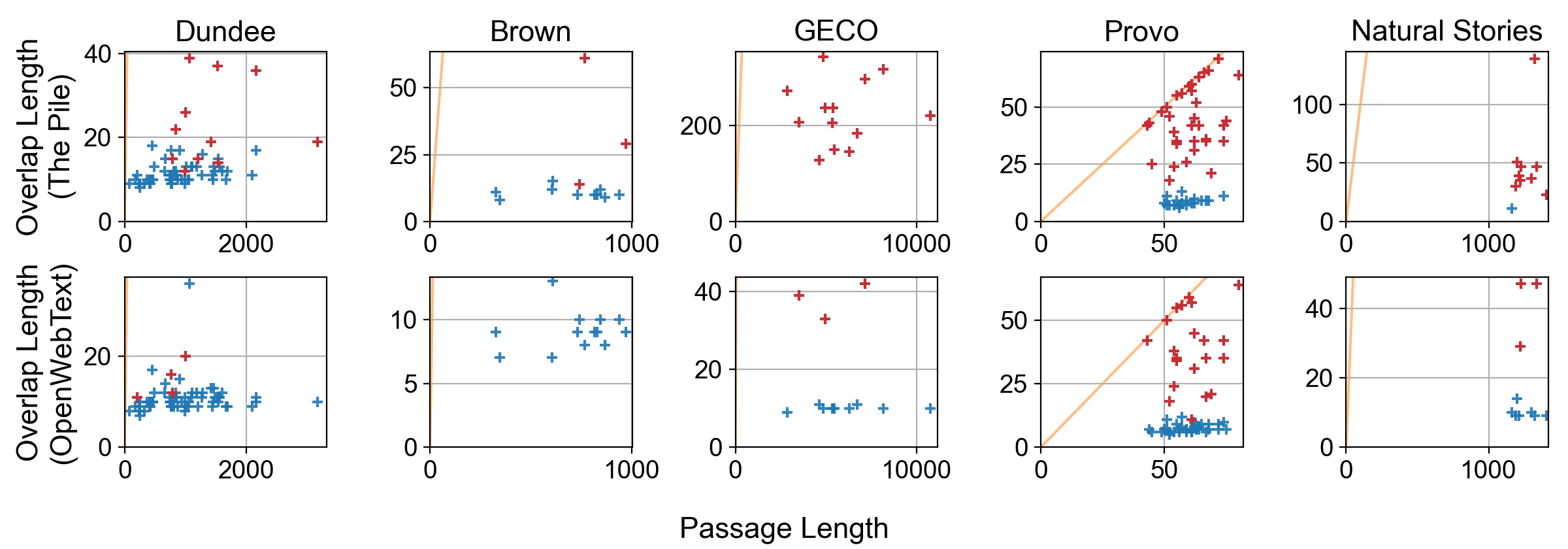}
        \caption{The length of each passage in the reading time corpora, and the length of its longest overlapping sequence with the Pile (top) and the OpenWebText Corpus (bottom), both measured in the number of subword tokens. The orange line denotes the $y=x$ line that indicates complete overlap.}
        \label{fig:overlap}
    \end{subfigure}
    \begin{subfigure}[t]{\textwidth}
        \includegraphics[width=\textwidth]{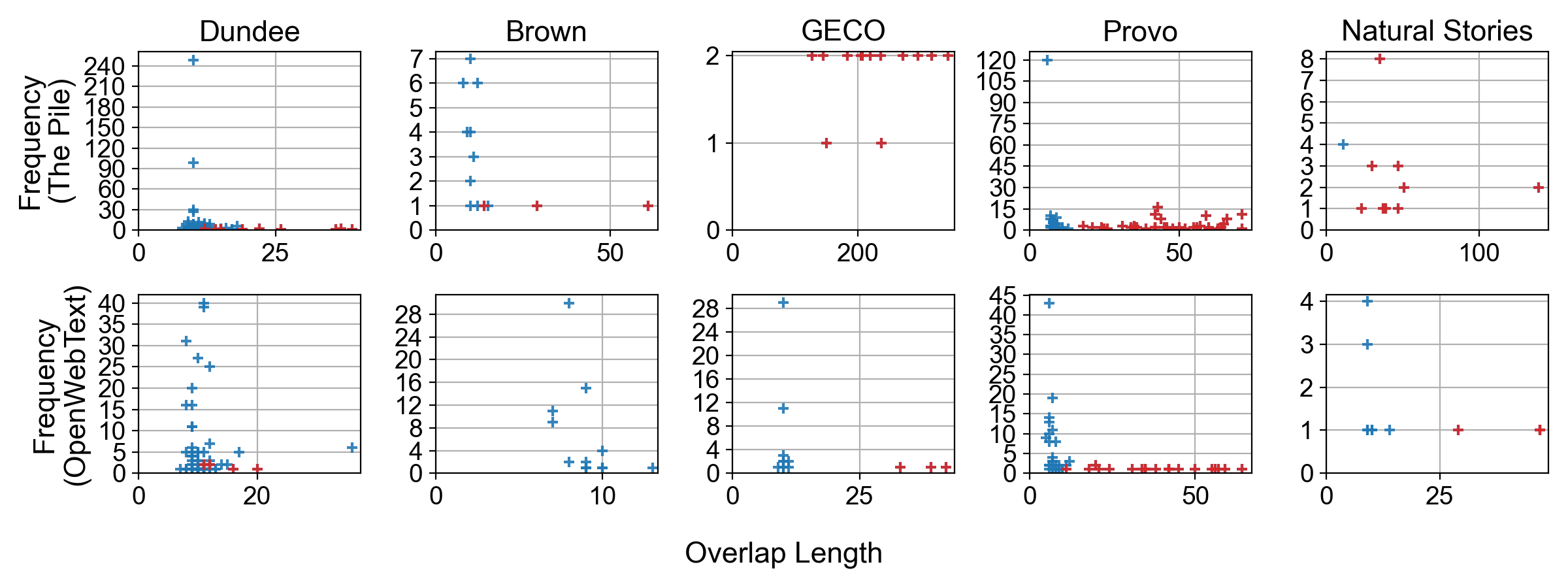}
        \caption{The length and frequency of the longest overlapping sequence between each passage of the reading time corpora and the Pile (top) and the OpenWebText Corpus (bottom). When there are multiple overlaps with the same maximum length, the highest frequency is reported.}
        \label{fig:frequency}
    \end{subfigure}
    \caption{Overlap between the five reading time corpora and the two pre-training corpora. Each `+' represents one passage, and the red `+'s denote longest overlapping sequences that have a probability lower than 0.05 to appear at least once in each corpus by chance (i.e.~sequences with 5-gram log probabilities lower than $-28.87$ and $-25.33$ for the Pile and OpenWebText respectively).}
\end{figure*}

While our method of detecting data leakage based on token sequence overlap is not robust against minor variations in surface form (e.g.~paraphrases), using a `softer' match criterion such as the similarity between sequence-level embeddings is computationally infeasible at the scale of the pre-training datasets we study, and may yield unreliable results depending on the quality of the embeddings.
Additionally, our method allows complete overlaps to be detected reliably, which similarity-based methods usually cannot allow when the lengths of the two sequences are different.

\subsection{Results} \label{sec:lm}
Figure \ref{fig:overlap} shows that except for the Provo Corpus, no passage in the reading time corpora is observed entirely in both pre-training datasets.
That is, the length of each passage's longest overlapping sequence is relatively short compared to the full passage length.
While the Provo passages that are observed in their entirety or the longer overlapping sequences in the Pile that exceed 100 tokens may especially be concerning, Figure \ref{fig:frequency} shows that such instances are very infrequent.
Most of the highly overlapping Provo passages each occur under 10 times in both pre-training corpora, and the longer overlapping sequences exceeding 100 tokens occur at most twice.
Therefore, we interpret these results as indicating that the reading time corpora suffer little from data leakage.\footnote{We publicly release the longest overlapping sequences and their frequencies at \url{https://github.com/byungdoh/rt-leakage}.}

\section{Study 2: The Influence of Leakage on Fit to Reading Times} \label{sec:exp2}
The previous study shows that most passages of the reading time corpora have not been leaked in the two pre-training corpora, which suggests that leakage is unlikely to be a possible explanation for the adverse effect of model size on LM surprisal's fit to reading times \citep{ohschuler23tacl, shainetal24}.
The second study causally verifies this by training LMs of different sizes on training examples that overlap minimally with the reading time corpora.
Additionally, to examine how surprisal's fit to reading times would change in the face of severe leakage, these LMs are fine-tuned on examples created from the reading time corpora.

\subsection{Methods} \label{sec:exp2_methods}
\paragraph{LM Training on Leakage-Free Data.}
We used the methodology of the previous study to identify training examples from the Pile that overlap minimally with the reading time corpora.
Specifically, CDAWGs were built separately on 143 chunks of 1,000 training batches.
A total of 18 chunks were found to have no more than 11 continuous tokens of overlap with any passage in the five reading time corpora, which excludes all overlaps improbable enough to meet our threshold in Figure \ref{fig:overlap}.
Among these, we sampled 10 chunks (i.e.~10,000 training batches of 1,024 examples; $\sim$20.9B tokens) as the training data.
One epoch of this `leakage-free' data was used to train Pythia-like Transformer LMs of three different sizes (Table \ref{tab:params}) using the GPT-NeoX library \citep{gptneox21}.\footnote{See Appendix \ref{sec:training} for additional training details.}

\begin{table}[t!]
    \centering
    \begin{tabular}{lrrrr} \toprule
    Model & \#L & \#H & $d_{\text{model}}$ & \#Parameters \\ \midrule
    \textit{Small} & 3 & 4 & 256 & 28,125,440 \\
    \textit{Medium} & 6 & 8 & 512 & 70,426,624 \\
    \textit{Large} & 12 & 12 & 768 & 162,322,944 \\ \bottomrule
    \end{tabular}
    \caption{Hyperparameters of LMs that were trained in this study. \#L: number of layers; \#H: number of attention heads per layer; $d_{\text{model}}$: embedding size.}
    \label{tab:params}
\end{table}

\begin{figure*}[ht!]
    \centering
    \includegraphics[width=\textwidth]{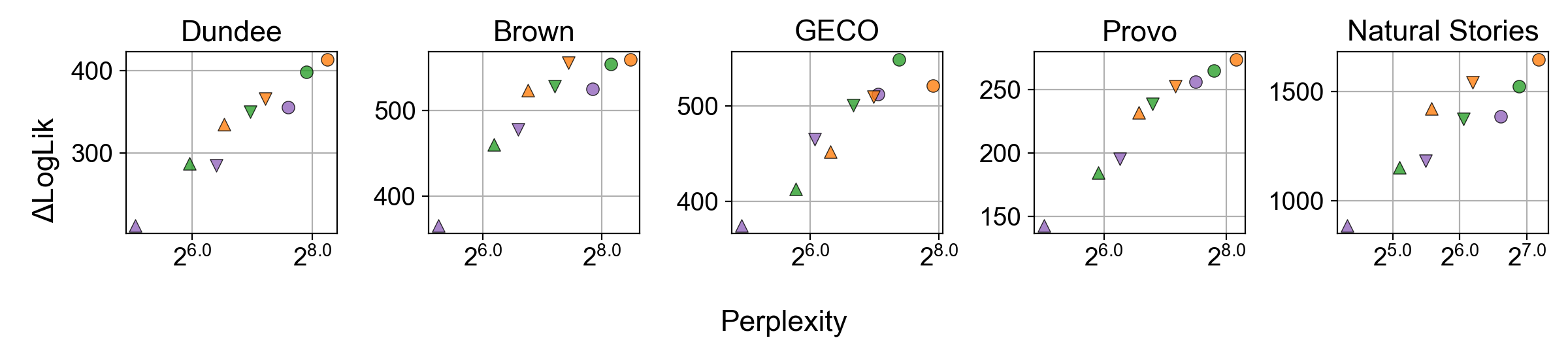}
    \includegraphics[width=0.495\textwidth]{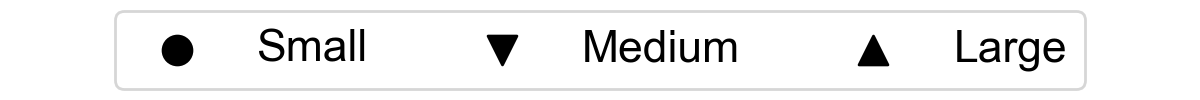}
    \includegraphics[width=0.495\textwidth]{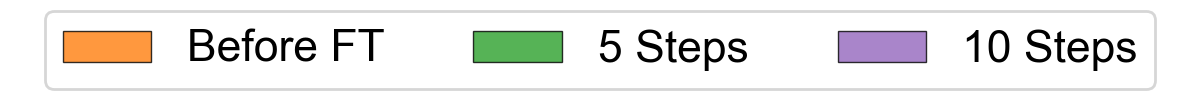}
    \caption{$\Delta$LogLik due to surprisal on held-out data and corpus-level perplexity from LMs trained leakage-free (orange) and after 5 and 10 fine-tuning (FT) steps on reading time data (green and purple respectively).}
\label{fig:delta_ll}
\end{figure*}

\paragraph{LM Fine-Tuning on Reading Time Data.}
After the LMs were trained, leakage was artificially introduced by fine-tuning them on examples created from the reading time corpora.
The construction procedure of the fine-tuning examples closely followed that of the Pythia training data.
First, the passages of the five reading time corpora were shuffled and concatenated with \texttt{<|endoftext|>} tokens inserted at passage boundaries to create one long sequence consisting of 165,643 tokens.
Subsequently, this sequence was split into contiguous sequences of length 2,048 to create one fine-tuning batch of 80 examples.
This procedure was repeated to generate additional batches, each containing the five reading time corpora, albeit in different order.
These batches were used to fine-tune each LM using the AdamW optimizer \citep{loshchilovhutter19} with a constant learning rate of 0.0001.
Results are reported after 5 and 10 fine-tuning steps.\footnote{We expect each fine-tuning step to serve as an upper bound for the effect of data leakage due to observing the same data during pre-training, given the recent and repeated nature of exposure during fine-tuning.}

\begin{table}[t!]
    \centering
    \begin{tabular}{lrr} \toprule
    Corpus/Measure & Fit & Exploratory \\ \midrule
    Brown SPR & 59,292 & 29,671 \\
    Natural Stories SPR & 384,984 & 192,826 \\ \midrule
    Dundee FP & 98,115 & 48,598 \\
    GECO FP & 144,850 & 72,468 \\
    Provo FP & 52,959 & 26,539 \\ \bottomrule
    \end{tabular}
    \caption{Number of data points in the fit and exploratory partitions of each reading time dataset.}
    \label{tab:observations}
\end{table}

\paragraph{Surprisal Calculation and Reading Time Modeling.}
The three LMs were used to calculate word-by-word surprisal on the five reading time corpora, after both initial training and subsequent fine-tuning.
When a passage did not fit into a context window of 2,048 tokens, the second half of the previous context window was used to condition the surprisal of the remaining tokens.
As the Pythia LM's subword tokens contain leading whitespaces, word probabilities were calculated with trailing whitespaces to ensure their consistency \citep{ohschuler24, pimentelmeister24}.\footnote{For example, without this correction, if both $\condprob{\texttt{\textvisiblespace car}}{\texttt{I \textvisiblespace sold \textvisiblespace the}}$ and $\condprob{\texttt{pet}}{\texttt{I \textvisiblespace sold \textvisiblespace the \textvisiblespace car}}$ are very high, the combined probabilities of ``\texttt{\textvisiblespace car}'' and ``\texttt{\textvisiblespace car pet}'' given the context ``\texttt{I \textvisiblespace sold \textvisiblespace the}'' can exceed one.}

Subsequently, for each LM, linear mixed-effects \citep[LME;][]{batesetal15} models that contain LM surprisal and standard baseline predictors were fit to about 50\% of the data points in each reading time dataset (fit partition; Table \ref{tab:observations}).\footnote{Self-paced reading times for Brown and Natural Stories, and first-pass durations for Dundee, GECO, and Provo. See Appendix \ref{sec:regression} for the full LME modeling details.}
The goodness-of-fit of each regression model was then evaluated by calculating the log-likelihood on about 25\% of held-out data points (exploratory partition).
This was compared against the log-likelihood of the baseline regression model that does not contain LM surprisal to evaluate the contribution of LM surprisal ($\Delta$LogLik).
All LME models incorporate maximal by-subject random effects \citep{barretal13} and assume a linear relationship between surprisal and reading times \citep{wilcoxetal23, xuetal23, shainetal24} and a lingering influence of surprisal from the previous word \citep{rayneretal83}.
The LMs' perplexity on each reading time corpus is also reported.

\subsection{Results}
Figure \ref{fig:delta_ll} shows that LMs trained on leakage-free data filtered according to this very strict criterion still demonstrate a negative relationship between model size and fit to reading times on all five datasets.
Together with the results of the previous study, this indicates that similar previous findings using Pythia and GPT-2 LMs \citep{ohschuler23emnlp, shainetal24} are not simply due to leakage.
The results from the LMs fine-tuned on reading time data in Figure \ref{fig:delta_ll} also show that if severe leakage were to exist, this would result in an overestimation of the strength of this negative relationship.
When examples from the reading time corpora are added to training, larger models seem to be able to predict certain words more accurately given the same number of fine-tuning updates, resulting in larger decreases in both perplexity and $\Delta$LogLik.
This suggests that smaller LMs are generally less susceptible to the influence of leakage, and that model-centered methods for diagnosing memorization \citep[e.g.~evaluating an LM's generated text given the prefix;][]{carlinietal23} may be effective for assessing leakage in very large LMs.

\section{Conclusion}
This study examines whether commonly used naturalistic reading time corpora have leaked into large-scale datasets on which LMs are trained.
In terms of sequence overlap, the leakage of most naturalistic reading time passages is found to be benign in two pre-training corpora.
While setting a criterion for what constitutes severe data leakage is difficult and requires some judgment, we hope that the overlapping sequences between commonly used reading time corpora and pre-training corpora identified in this work provide a resource for those that wish to be more careful with psycholinguistic modeling.

The subsequent regression experiment replicates the negative relationship between model size and surprisal's fit to reading times using LMs trained on leakage-free data.
Taken together, these results suggest that previously reported findings using LMs trained on these corpora are not driven by the effects of data leakage.
In contrast to \citet{wilcoxetal23qp}, who analyzed the potential influence of data leakage using smaller LMs trained from scratch, these results provide more direct evidence that generalizes to trends observed from larger pre-trained LMs like Pythia and GPT-2.

Previous studies have shown that more accurate predictions of named entities and other low-frequency words primarily drive this inverse scaling trend of pre-trained LM surprisal \citep{ohschuler23tacl, ohetal24}.
This study additionally suggests that the ability of larger models to predict such words more accurately is not due to `memorizing' exact sequences, but rather due to more complex word-to-word associations learned during training.

\section*{Acknowledgments}
We thank the ARR reviewers and the area chair for their helpful comments, and William Merrill for his help with CDAWG construction.
This work was supported by the National Science Foundation (NSF) grant \#1816891.
All views expressed are those of the authors.
This work was supported in part through the NYU IT High Performance Computing resources, services, and staff expertise.

\section*{Limitations}
In this work, the potential leakage of naturalistic text stimuli is evaluated through studies using English corpora, language models trained on English text, and reading time data from native speakers of English.
Therefore, replication studies are necessary to further assess the leakage of text stimuli in other languages.
Additionally, data leakage in this work is diagnosed mainly through token $n$-gram overlaps, which is insensitive to minor variations in form.
Moreover, as the OpenWebText Corpus is an open-source effort to replicate GPT-2's undisclosed training data, the corpus statistics of the actual training data may differ.
Finally, this work is concerned with the use of language models as cognitive models of human sentence processing, and therefore does not relate to their use in natural language processing applications.

\section*{Ethics Statement}
This work used publicly available text corpora \citep{gokaslancohen19, gaoetal20} and human subject data collected as part of previously published research \citep{futrelletal21, smithlevy13, copetal17, kennedyetal03, lukechristianson18}.
Readers are referred to the respective publications for more information about the data collection and validation procedures.
As this work studies the connection between language models and human sentence processing, its potential negative impacts on society appear to be minimal.

\bibliography{custom}

\appendix

\begin{table*}[ht!]
    \centering
    \begin{tabular}{lr} \toprule
    Datasets & LME Formula \\ \midrule
    \makecell[l]{Brown \\ Natural Stories} & \makecell[r]{\texttt{RT $\sim$ LMsurp + LMsurp\_prev + Unisurp + length + index +} \\ \texttt{(LMsurp + LMsurp\_prev + length + index + 1 | subject)}} \\ \midrule
    \makecell[l]{Dundee \\ GECO \\ Provo} & \makecell[r]{\texttt{RT $\sim$ LMsurp + LMsurp\_prev + Unisurp + length + index + pfix +} \\ \texttt{(LMsurp + index + 1 | subject)}} \\ \bottomrule
    \end{tabular}
    \caption{Formulae of LME models fit in Study 2. \texttt{LMsurp}: LM surprisal, \texttt{LMsurp\_prev}: LM surprisal of previous word, \texttt{Unisurp}: unigram surprisal, \texttt{length}: word length, \texttt{index}: position of the word within the sentence, \texttt{pfix}: whether the previous word was fixated. The baseline regression models were fit with these formulae without the \texttt{LMsurp} and \texttt{LMsurp\_prev} predictors. All predictors were z-transformed.}
    \label{tab:formula}
\end{table*}

\section{LM Training Details}
\label{sec:training}
The LMs in Study 2 were trained closely following the training procedures of the Pythia LMs.
Like many other Transformer LMs, Pythia LMs are based on decoder-only layers, but they parallelize the computations of the attention mechanism and those of the feedforward neural network, and do not use tied parameters for the input embedding and final projection matrices.
The three LMs were trained using the Zero Redundancy Optimizer \citep[ZeRO;][]{rajbhandarietal20} implementation of Adam \citep{kingmaba15} with a maximum learning rate of 0.001.
This learning rate was warmed up linearly over the first 1\% of training steps (i.e.~100 steps) and was subsequently lowered to a minimum of 0.0001 following a cosine annealing schedule over the remainder of the 10,000 training steps.
Gradients were clipped to a maximum norm of 1 prior to each update to stabilize training.
All training took place in half-precision on 48GB Nvidia RTX 8000 GPUs.

\section{LME Modeling Details}
\label{sec:regression}
\paragraph{Data Preprocessing and Partitioning.}
For the Brown and Natural Stories datasets, reading times of words at sentence boundaries and those shorter than 100 ms or longer than 3,000 ms were excluded.
Data from subjects who answered four or fewer comprehension questions correctly were also removed from the Natural Stories data.
The Dundee, GECO, and Provo datasets were filtered to exclude reading times of unfixated words, words following saccades longer than four words, and words at sentence and document boundaries.
Reading times of words at line and screen boundaries were also removed from the Dundee data that provides annotations of line/screen locations.

After data preprocessing, each dataset was partitioned into fit and exploratory partitions that comprise roughly of 50\% and 25\% of the data respectively (Table \ref{tab:observations}).
This partitioning was based on the sum of the subject ID and the sentence ID, which keeps all data from a particular subject-sentence combination intact in one partition.
Each fit partition was used to fit the regression models, and the exploratory partition was used to calculate regression model likelihood.
The remaining $\sim$25\% of the data is reserved for statistical significance testing and was not used in this work.

\paragraph{LME Model Specifications.}
The baseline predictors included in all LME models are word length in characters, index of word position within the sentence, unigram surprisal (all datasets), and whether the previous word was fixated (Dundee, GECO, Provo only).
Unigram surprisal was calculated using the KenLM toolkit \citep{heafieldetal13} with probabilities estimated on the OpenWebText Corpus \citep{gokaslancohen19}.
The by-subject random effects structures of the LME models were determined by starting with maximal random effects and removing the least predictive random effect until all LME models converged.
The resulting LME formulae are outlined in Table \ref{tab:formula}.

\end{document}